\documentclass[conference]{IEEEtran}
\IEEEoverridecommandlockouts
\usepackage{booktabs}
\usepackage[numbers,sort&compress]{natbib}
\usepackage{amsmath,amssymb,amsfonts}
\usepackage{algorithmic}
\usepackage{multirow}
\usepackage{graphicx}
\usepackage{textcomp}
\usepackage{xcolor}
\usepackage{adjustbox}
\usepackage{ulem} 
\usepackage[margin=1in]{geometry} 

\def\BibTeX{{\rm B\kern-.05em{\sc i\kern-.025em b}\kern-.08em
    T\kern-.1667em\lower.7ex\hbox{E}\kern-.125emX}}
\begin{document}

\title{Predicting Rental Price of Lane Houses in Shanghai with Machine Learning Methods and Large Language Models\\
}

\author{\IEEEauthorblockN{Tingting Chen}
\IEEEauthorblockA{\textit{Department of Data Science and Big Data Technology} \\
\textit{Shanghai International Studies University}\\
Shanghai, China \\
18790061936@163.com}
\and
\IEEEauthorblockN{Shijing Si$^{\star}$\thanks{$^\star$Corresponding author: shijing.si@shisu.edu.cn}}
\IEEEauthorblockA{\textit{Department of Data Science and Big Data Technology} \\
\textit{Shanghai International Studies University}\\
Shanghai, China \\
shijing.si@shisu.edu.cn}
}

\maketitle

\begin{abstract}
Housing has emerged as a crucial concern among young individuals residing in major cities, including Shanghai. Given the unprecedented surge in property prices in this metropolis, young people have increasingly resorted to the rental market to address their housing needs.
This study utilizes five traditional machine learning methods—multiple linear regression (MLR), ridge regression (RR), lasso regression (LR), decision tree (DT), and random forest (RF)—along with a Large Language Model (LLM) approach using ChatGPT, for predicting the rental prices of lane houses in Shanghai. It applies these methods to examine a public data sample of about 2,609 lane house rental transactions in 2021 in Shanghai, and then compares the results of these methods. In terms of predictive power, RF has achieved the best performance among the traditional methods. However, the LLM approach, particularly in the 10-shot scenario, shows promising results that surpass traditional methods in terms of R-Squared value. The three performance metrics—mean squared error (MSE), mean absolute error (MAE), and R-Squared—are used to evaluate the models. Our conclusion is that while traditional machine learning models offer robust techniques for rental price prediction, the integration of LLM such as ChatGPT holds significant potential for enhancing predictive accuracy.
\end{abstract}


\section{Introduction}

Housing plays a vital role in our lives, particularly in cities like Shanghai where high housing prices present significant challenges for tenants. The considerable variation in rental prices for lane houses adds complexity to housing choices, especially with the evolving demands for living environments driven by rapid urbanization.

Understanding the factors influencing rental prices encompasses several dimensions. Firstly, the scarcity of land resources and housing due to urbanization significantly contributes to high rental prices of lane houses\cite{wu2001china}. Urban development constraints and limited mobility of lane houses create ample opportunity for rental price escalation. Secondly, changing demands for living environments play a crucial role. While traditional lane houses hold cultural significance, issues like limited space, short duration of use, and outdated facilities gradually diminish their appeal\cite{chen2020measuring}.

Moreover, social and policy factors also influence rental prices. Land use restrictions in Shanghai's core areas limit the number of lane houses, leading to higher rental prices and relative disadvantages for residents. Government policies on rental and transportation further impact rental prices. For instance, public transportation development and property rental regulations significantly shape urban residents' residential choices\cite{li2019analyzing}.

Studying rental price determinants is valuable for tenants, facilitating more accurate rental estimates and ensuring market stability. Additionally, it contributes to urbanization, social progress, and improved quality of life. For residents in lane houses, understanding rental prices expands housing options and enhances quality of life.

To address this, we propose employing a combination of traditional machine learning methods and advanced Large Language Models (LLM) to predict lane house rental prices in Shanghai. The traditional machine learning methods include multiple linear regression, ridge regression, lasso regression, decision tree, and random forest. Additionally, we explore the use of ChatGPT, a state-of-the-art LLM, in various shot scenarios (0-shot, 1-shot, 5-shot, and 10-shot) to assess its performance in predicting rental prices. Utilizing relevant features, we estimate prices and compare results to identify the most effective algorithm. Our main work is summarized as follows:

\begin{itemize}
\item First, we took the Kaggle dataset on Shanghai lane house rentals and identified the important features that accurately reflect the rental prices.

\item Next, we preprocessed the dataset, selected the variables that are related to predicting rental prices and analyzed the dataset by visualization.

\item Third, we applied five traditional machine learning methods including Multiple Linear Regression, Ridge Regression, Lasso Regression, Decision Tree, and Random Forest to predict the house price respectively.

\item Fourth, we implemented ChatGPT in 0-shot, 1-shot, 5-shot, and 10-shot scenarios to predict the rental prices, utilizing its ability to process natural language prompts and generate predictions based on contextual information.

\item Lastly, by comparing their outputs such as MSE, MAE, and R Squared, we determined the optimal model and assessed the potential of LLM in enhancing predictive accuracy.

\end{itemize}

Through this approach, we aim to provide a comprehensive comparison between traditional machine learning models and LLM, highlighting the strengths and limitations of each method in the context of rental price prediction.

\section{Related Work}
This paper is related to two lines of research: current research status on rental houses and machine learning methods.

\subsection{Current Research Status on Rental Houses}
Over the past three decades, China's real estate market has undergone significant changes due to its transition to a global market economy \cite{zhou2023market}. The government has shifted its focus from speculation to long-term housing mechanisms, emphasizing both renting and purchasing options. This reform direction indicates a return to housing properties. The long-term rental apartment market experienced substantial investment expansion from 2017 to 2018, driven by government policies aimed at regulating the real estate market \cite{clark2021can}. Overall, China's institutional framework and government policies are facilitating the expansion of the rental market.

Based on tenants' needs, various factors can influence their choice of rental houses, including location, amenities, price, and more. Zhou identified commuting distance as a significant factor influencing housing decisions. And he developed a Cobb-Douglas model utilizing housing area and commuting distance as primary monitoring variables, highlighting the importance of commuting distance, community environment, and allocation level in housing decision-making \cite{zhou2016impact}. This model comprehensively addresses residents' needs and holds significant implications for future research. 
And Zheng explored factors influencing rental housing among the young generation in Chinese cities using the theory of planned behavior (TPB). Through structural equation model validation, they found that attitudes towards behavior, mandatory policies, and government incentives significantly influence renting behavior \cite{zheng2019understanding}. 
At the same time, the Chinese government has gradually paid attention to the development of the rental housing market and began to issue a large number of relevant policies by the central and local governments in recent years \cite{guo2024policy}.

\subsection{Machine Learning Methods}
There has been a surge in utilizing machine learning techniques for predicting housing prices in recent years. And this trend is driven by the increasing complexity of housing markets and the need for more accurate prediction models. 

In the pursuit of predicting rental prices for lane houses in Shanghai, researchers have explored various modeling techniques. While traditional economic models such as the Hedonic Price Model (HPM) have been employed by some\cite{cebula2009hedonic} \cite{wen2005hedonic}\cite{dunse1998hedonic}\cite{mok1995hedonic}, they often struggle to capture the complex non-linear relationships inherent in unique housing types, particularly in urban centers. 

For instance, Madhuri\cite{ghosalkar2018real} employed linear regression to predict residential prices in Mumbai, demonstrating minimal prediction errors and highlighting the efficacy of linear regression. Similarly, Zhou\cite{zhou2021house} integrated empirical analysis of the Washington DC real estate market with theoretical housing price models, employing multiple regression with particle swarm optimization to improve prediction accuracy. Additionally, Zaki et al.'s research\cite{zaki2022house} aimed to predict house prices, comparing XGBoost with traditional hedonic regression pricing. Their study underscored the practicality of XGBoost in predicting house prices, reporting significantly higher accuracy compared to hedonic regression.

In addition to linear regression, researchers often employ Ridge and Lasso regression in conjunction with linear regression for house price prediction to reduce errors. Rh et al.\cite{rh2022price} developed models combining linear regression with k-nearest neighbors algorithm and gradient descent optimization, integrating personalized housing preferences based on faith and budget considerations. Sharma et al.\cite{sharma2024house} explore house price prediction in Bangalore using linear, Lasso, and Ridge regression, incorporating factors such as area, number of bathrooms, and climate conditions to enhance accuracy. Moreover, Manasa\cite{manasa2020machine} constructs a predictive model for Bengaluru's house prices using multiple linear regression, Lasso, and Ridge regression, alongside Extreme Gradient Boost Regression (XG Boost). The study emphasizes error analysis to identify optimal predictive approaches, including the utilization of mixed models.

The Decision Tree model is widely applied. Zhang\cite{zhang2021decision} proposes an objective scheme using a decision tree, outperforming other machine learning methods by emphasizing crucial features such as housing density, population quality, location, educational resources, and crime rates. Kuvalekar\cite{kuvalekar2020house} forecasts house prices in Mumbai's real estate market using a Decision Tree Regressor, achieving impressive accuracy and aiming to enhance investment decision-making. Additionally, Yang \cite{yang2021machine} employs gradient boosting decision trees (GBDT) to analyze the non-linear relationship between BRT and house prices in Xiamen, highlighting GBDT's superiority over traditional econometric techniques.

Several studies have explored the use of Random Forest algorithms for predicting property prices in different regions. Tanamal\cite{tanamal2023house} investigates property value forecasting in Surabaya, integrating real estate agents' insights into the model, achieving an 88\% accuracy rate. Jui et al.\cite{jui2020flat} propose a flat price prediction framework for Dhaka, Bangladesh, demonstrating the superiority of random forest regression over linear regression. Adetunji\cite{adetunji2022house} emphasizes the limitations of traditional methods like the House Price Index, advocating for the application of Random Forest for house price prediction.

While the machine learning methods mentioned above each have their own advantages and disadvantages, there remains a lack of comprehensive studies comparing these methods. Our work aims to fill this gap by employing various machine learning methods to predict rental lane house prices and systematically comparing their results.

\subsection{Large Language Models in Predictive Tasks}
Recently, large language models (LLM) such as GPT-3, ChatGPT and BERT have demonstrated remarkable capabilities in a wide range of natural language processing tasks, including predictive modeling. LLM leverage vast amounts of data and complex neural network architectures to understand and generate human-like text, making them highly versatile tools.

LLM have been successfully applied to various predictive tasks beyond traditional NLP applications. For instance, Brown et al.\cite{brown2020language} demonstrated that GPT-3 could generate coherent text, perform arithmetic, translate languages, and even solve complex problems with minimal input. The adaptability of LLM to different contexts makes them promising candidates for predictive modeling in domains such as finance, healthcare, and real estate.

In the context of real estate, LLM can be used to predict housing prices by processing and analyzing vast amounts of textual data related to property descriptions, market trends, and economic indicators. For example, Heidari and Rafatirad\cite{heidari2020bidirectional} presented a model using Bidirectional Encoder Representations from Transformers (BERT) for predicting rent prices based on online information from various real estate websites, illustrating the effectiveness of LLM in handling diverse data sources.

And our study also explores the application of ChatGPT in predicting lane house rental prices in Shanghai, comparing its performance in various shot scenarios with traditional machine learning methods.


\section{DATA PREPROCESSING}
"Shanghai Lane House Rental Prices 2021" dataset, which contains over 2000 rental price records and their associated property variables. These variables, functioning as features within the dataset, were subsequently employed to predict the average price per square meter for each house. Below are a few feature engineering processes which were done to clean the dataset:

\begin{itemize}
\item 1. Remove missing values, and the dataset decreased from 2608 to 2607 entries.
\item 2. Eliminate duplicate values, and the dataset decreased from 2607 to 2549 entries.
\item 3. Irrelevant factors with minimal impact on rent were removed using an Excel spreadsheet.
\item 4. Add a new variable called "total-ssvalue" to facilitate subsequent modeling. (This variable represents the cumulative count of various soft furnishing facilities such as air conditioning, heating, and other amenities.)
\end{itemize}
\vspace{1em} 

After basic data preprocessing, here is a dataset containing 16 attributes that can be used to predict rental prices in Shanghai's "longtang" houses. These attributes include:
\begin{itemize}
\item 1. Area: The district where the house is located.
\item 2. Rental Price: The amount of rent for the house.
\item 3. Number of Bedrooms: The number of bedrooms in the house.
\item 4. Living and Dining Rooms: Whether the house has separate living and dining rooms.
\item 5. Number of Bathrooms: The number of bathrooms in the house.
\item 6. Loft: Whether the house has a loft.
\item 7. Square Meters: The size of the house in square meters.
\item 8. Heating Method: The heating method for the house.
\item 9. Air Conditioning: Whether the house has air conditioning.
\item 10. Balcony: Whether the house has a balcony.
\item 11. WIFI: Whether the house provides WIFI network.
\item 12. Outdoor Space: Whether the house has outdoor recreational space.
\item 13. Bathtub: Whether the house has a bathtub.
\item 14. Floor Heating: Whether the house has floor heating.
\item 15. Oven: Whether the house has an oven.
\item 16. All Facilities: All facilities and conveniences provided by the house.
\end{itemize}


Next, we will initiate the quantization process for the sample data. Quantization entails converting continuous data into discrete data. This enables a more comprehensive understanding and analysis of the data, thereby providing stronger support for modeling and prediction.

\subsection{Quantization of ordinal variables}\label{AA}
When dealing with variables exhibiting a hierarchical structure, employing the rank assignment method proves effective in capturing the inherent characteristics and their impact on apartment rent\cite{budiman2023machine}\cite{mccord2020house}. Take the total number of furnishings in a room, for instance, which can only be qualitatively described. Using rank assignment, it can be categorized into three types: [6,8] (ranked as 3), [3,6) (ranked as 2), and [0,3) (ranked as 1). This method helps depict and characterize room facilities. A greater variety of facilities enhances the rental experience, prompting tenants to be more willing to pay higher prices. Consequently, apartment operators typically price units with abundant facilities at a premium.

\subsection{Quantification of dummy variables}
\begin{itemize}
\item There is no hierarchical relationship among variable categories, and they are mutually exclusive. Consequently, feature variables like balcony or loft presence require virtual quantization for enhanced analysis \cite{xu2021new}\cite{liu2022second}. Variables such as room types with balconies or air conditioning, impacting lighting and ventilation, are distinctly quantized to better reflect their influence on prospective tenants' preferences and length of stay.

\item The variable "district" lacks a numerical representation but still influences rent. To align the data with algorithms and libraries, one-hot encoding is necessary \cite{avanijaa2021prediction} \cite{sinha2020enhanced}. This process involves converting textual data into numerical values.
\end{itemize}

 We divide the influencing factors into three main categories: regional features, room type features, and soft furnishing facilities. After data processing, the regional features consist of 14 variables (district), the room type features include the number of bedrooms (bedrooms), living and dining areas (living-dining), bathrooms, loft, area in square meters (sqmeters), building-type, and usage type (use-type-en), and the soft furnishing feature includes the total number of soft furnishing facilities (total-ssvalue). In total, there are 22 variables in TABLE \uppercase\expandafter{\romannumeral1}.

\begin{table}[h]
\centering
\begin{minipage}{0.5\textwidth}
\caption{List of Attributes} 
\resizebox{\textwidth}{!}{%
\begin{tabular}{lll}
\hline
\textbf{Attribute Name} & \textbf{Data Type} & \textbf{Description} \\
\hline
district & object & The district where the house is located \\
rent  & float64 & The amount of rent for the house \\
bedrooms & int64 & The number of bedrooms in the house \\
living-dining & int64 & Whether the house has separate living and dining rooms \\
bathrooms & int64 & The number of bathrooms in the house \\
loft & int64 & Whether the house has a loft \\
sqmeters & int64 & The size of the house in square meters \\
building-type & int64 & The type of the rental houses \\
use-type-en & int64 & Intended use of the rental houses \\
total-ssvalue & int64 & Total facilities of the rental houses \\
district-Baoshan & bool & The rental houses in Baoshan \\
district-Changning & bool & The rental houses in Changning \\
district-Hongkou & bool & The rental houses in Baoshan \\
district-Huangpu & bool & The rental houses in Huangpu \\
district-Jiading & bool & The rental houses in Jiading \\
district-Jing'an & bool & The rental houses in Jing'an \\
district-Minhang & bool & The rental houses in Minhang \\
district-Pudong & bool & The rental houses in Pudong \\
district-Putuo & bool & The rental houses in Putuo \\
district-Qingpu & bool & The rental houses in Qingpu \\
district-Songjiang & bool & The rental houses in Songjiang \\
district-Xuhui & bool & The rental houses in Xuhui \\
district-Yangpu & bool & The rental houses in Yangpu \\
district-Zhabei & bool & The rental houses in Zhabei \\
\hline
\end{tabular}
}
\end{minipage}
\end{table}

\section{Data Analysis}
After preprocessing the data, it undergoes data analysis to gain insights to predict rental prices for Shanghai lane houses. We employ pie charts, box plots, and heatmaps to perform detailed analysis, examining how each feature varies and its relationship with other features, including the target variable and rental prices.

Fig. 1 displays the distribution of rental properties in different districts of Shanghai in the first half of 2021. It is evident that Xuhui, Huangpu, Jing'an, and Changning districts have a larger number of available rental properties, making them the hotspots in the rental market.

\begin{figure*}[htbp]
\centerline{\includegraphics[width=\linewidth]{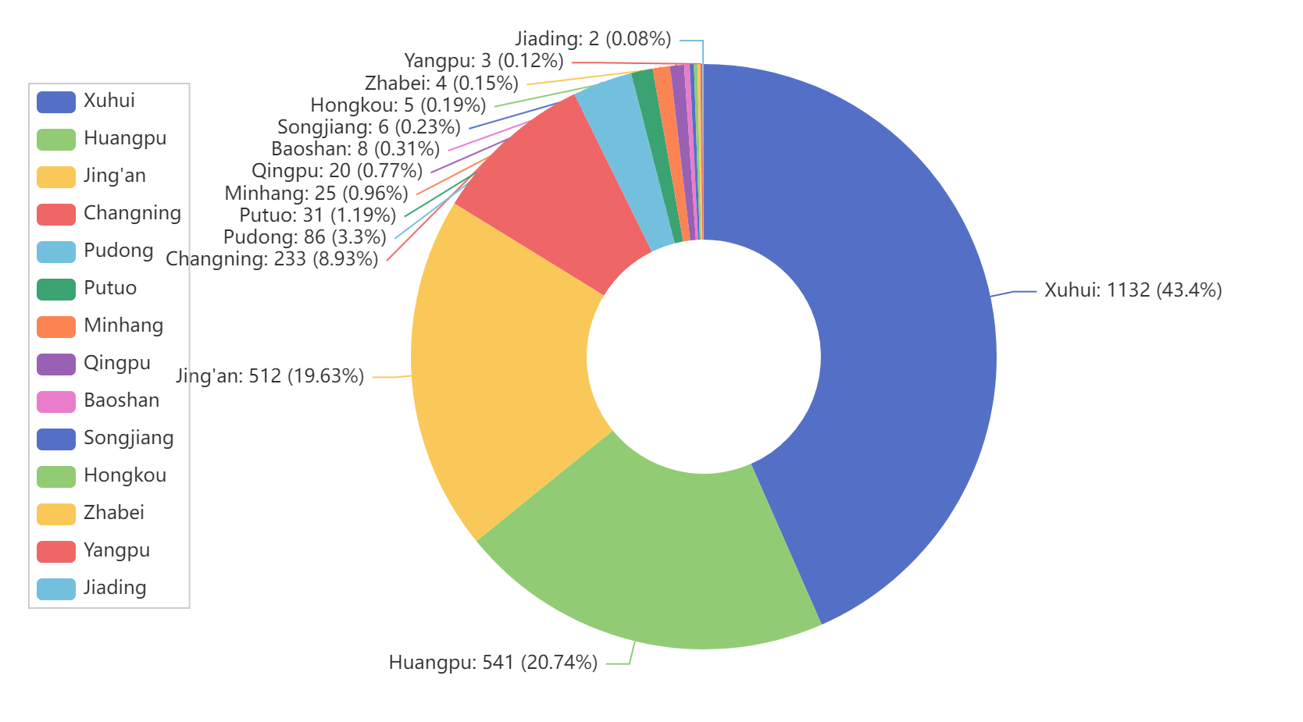}}
\caption{Distribution of rental properties in different regions.}
\label{fig}
\end{figure*}

Fig. 2 presents the distribution of rental prices in different regions. From the box plot, it can be observed that the region has a significant impact on the rental prices. Notably, areas such as Pudong, Minhang, and Qingpu exhibit noticeably higher rental prices. Surprisingly, even the suburban areas of Qingpu and Songjiang show relatively high rental levels.

\begin{figure*}[htbp]
\centerline{\includegraphics[width=\linewidth]{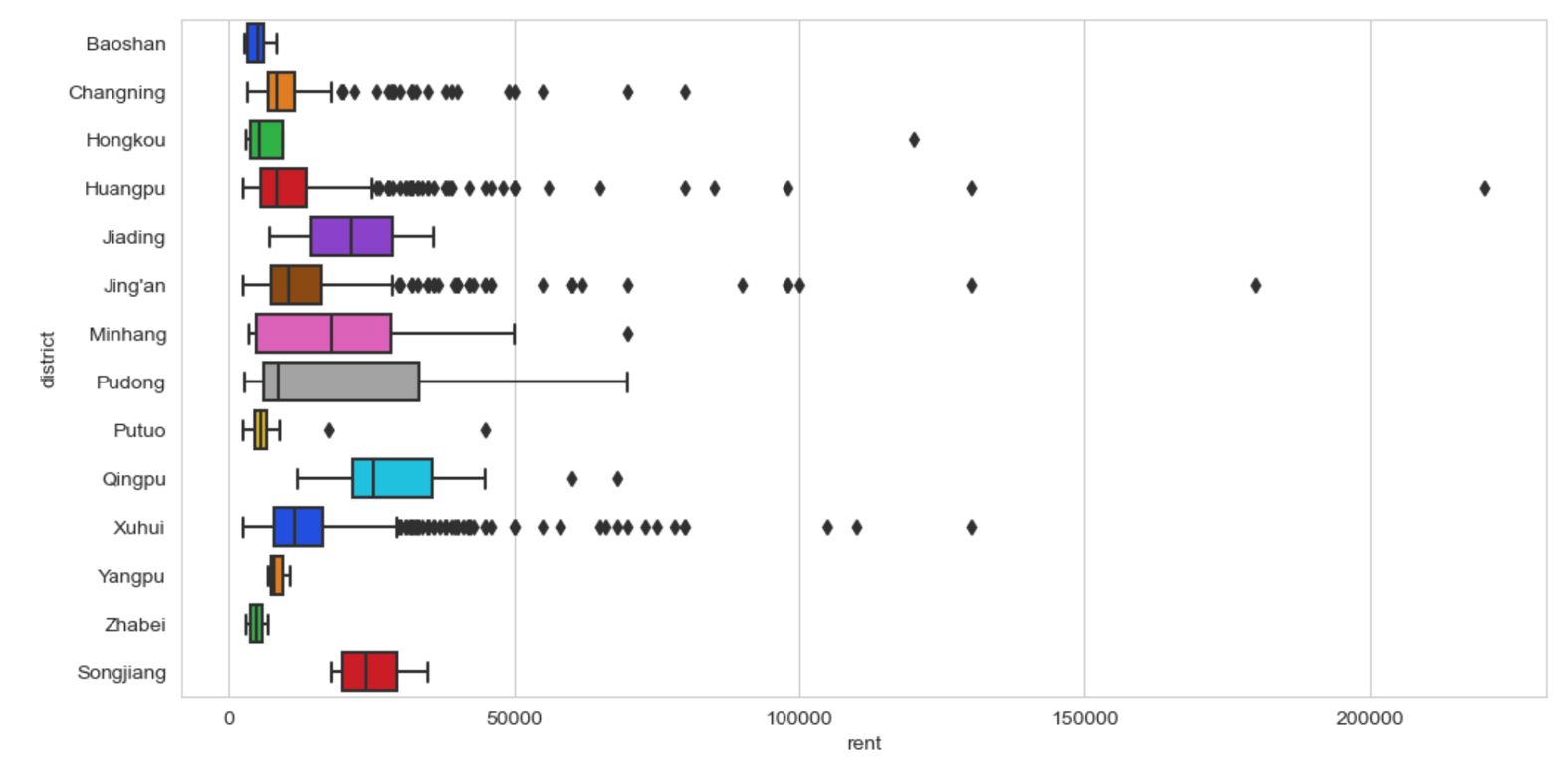}}
\caption{distribution of rental prices in different regions.}
\label{fig}
\end{figure*}

Fig. 3 displays the distribution of rental property sizes in different regions. By examining the pie chart, we can observe that areas like Xuhui, Huangpu, Jing'an, and Changning have larger average rental property sizes, while the suburban areas have smaller average rental property sizes.

\begin{figure*}[htbp]
\centerline{\includegraphics[width=\linewidth]{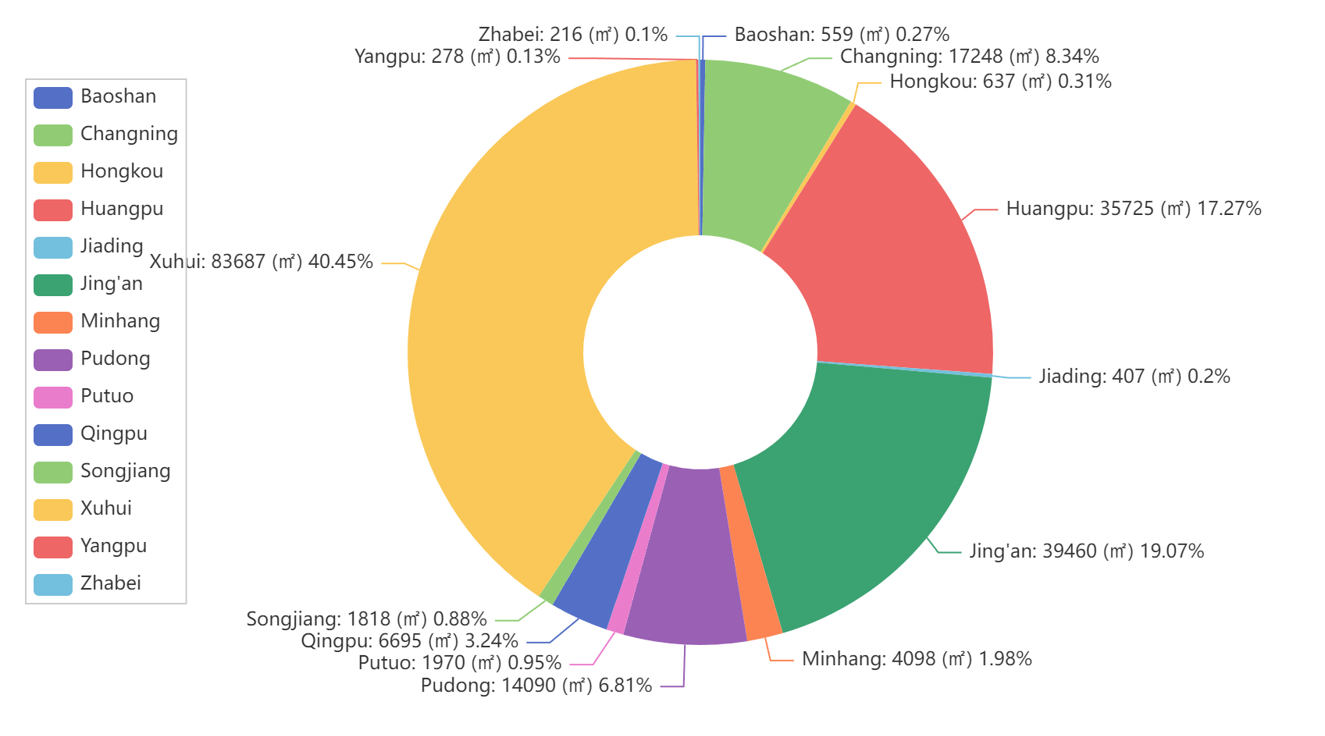}}
\caption{Distribution of property sizes in different regions.}
\label{fig}
\end{figure*}

Fig. 4 showcases the distribution of total soft furnishing facilities in rental properties across different regions. From the heatmap, we can clearly observe that Xuhui, Huangpu, Jing'an, and Changning areas have a significant abundance of soft furnishing facility types in rental properties.

\begin{figure*}[htbp]
\centerline{\includegraphics[width=\linewidth]{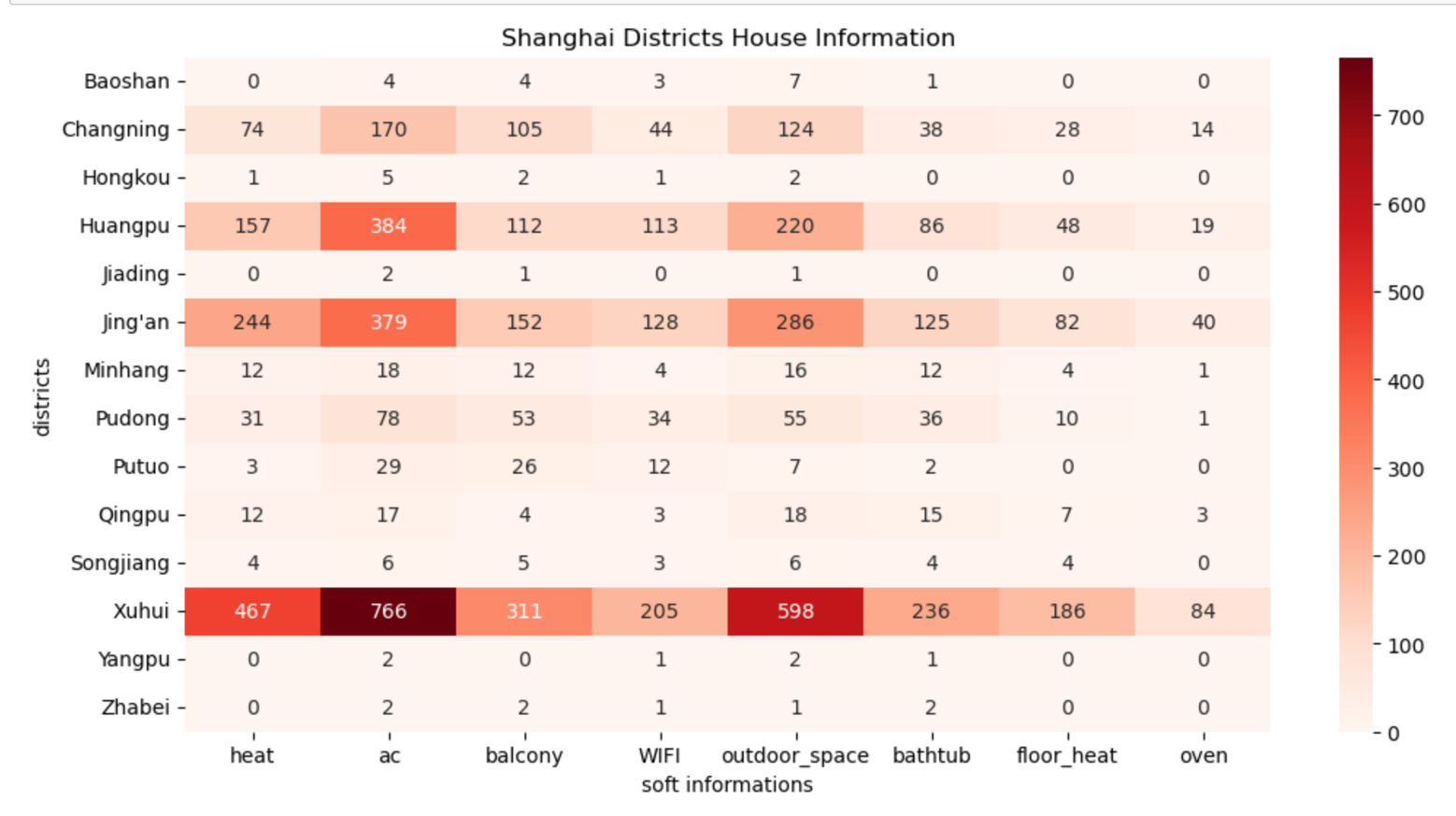}}
\caption{distribution of total soft furnishing facilities.}
\label{fig}
\end{figure*}

\section{MACHINE LEARNING MODEL SELECTION}
After we got the processed data, then, the dataset was split into training set and test set with a ratio of 4 : 1 by utilizing the scikit-learn package. Evaluation function used in this study is mean squared error (MSE), mean absolute error (MAE) and R Squared\cite{chicco2021coefficient}. This function is illustrated as follow:

\begin{align}
\text{MSE} &= \frac{1}{N} \sum_{i=1}^{N} (y_i - \hat{y}_i)^2 \\
\text{MAE} &= \frac{1}{N} \sum_{i=1}^{N} |y_i - \hat{y}_i| \\
R^2 &= 1 - \frac{\sum_{i=1}^{N} (y_i - \hat{y}_i)^2}{\sum_{i=1}^{N} (y_i - \bar{y})^2}
\end{align}

\subsection{Multiple Liner Regression}\label{AA}
Multiple Linear Regression (MLR)\cite{uyanik2013study} is the most common form of linear regression, serving as a fundamental tool for forecasting. MLR illustrates the correlation between a continuous dependent variable and two or more independent variables. This relationship is captured through equations such as Eq. 4 and Eq. 5:

\begin{equation}
E(Y |X) = \alpha_1 + \beta_1X_1 + \ldots + \beta_pX_p 
\end{equation}

where \(\alpha_1\) represents the intercept, and \(\beta_j\) are coefficients or slopes. By analyzing how responses vary around their respective mean values, we can further understand the model:

\begin{equation}
Y_j = \alpha_1 + \beta_1X_{j,1} + \ldots + \beta_pX_{j,p} + \epsilon_j 
\end{equation}

This formulation, equivalent to:

\begin{equation}
Y_j = E(Y |X_j) + \epsilon_j
\end{equation}

where \(Y_j\) represents the actual value and \(\epsilon_j\) denotes the error term, provides insight into the variability of responses. Just like XGBoost, MLR is accessible as an open-source package and has demonstrated its effectiveness in various machine learning and data mining challenges.




\subsection{Ridge Regression}
Ridge Regression (RR), serves as a valuable tool for analyzing multiple regression when dealing with multicollinearity (mcl) in the dataset. Multicollinearity, the presence of near-linear relationships among independent variables, can pose challenges to traditional regression models. Ridge Regression\cite{dorugade2014new} introduces a special condition on parameters, as expressed in Eq. 8 and Eq. 9:
\begin{equation}
\hat{\beta}_{\text{ridge}} = \arg\min_{\beta} \left\{ \sum_{a=1} (y_a - \sum_{b=1} x_{ab}\beta_b)^2 + \lambda \sum_{b=1} \beta_b^2 \right\} 
\end{equation}

This is equivalent to minimizing:

\begin{equation}
\sum_{i=1} \left( y_i - \sum_{j=1} x_{ij}\beta_j \right)^2 \bigg/ \sum_{i=1} \left( y_i - \sum_{j=1} x_{ij}\beta_j \right)^2 
\end{equation}

Subject to the constraint, for some \(c > 0\):

\begin{equation}
\sum_{j=1} \beta_j^2 < c \sum_{j=1} \beta_j^2 < c
\end{equation}

This constraint effectively controls the sum of the squared coefficients.


\subsection{Lasso Regression}
Lasso Regression (LR)\cite{reid2016study} is another variant of linear regression that utilizes a shrinkage technique, as described in Eq. 10. In Lasso Regression, the positive regularization parameter $\lambda_1$ plays a crucial role in controlling the values of the parameters $\beta$ and $\beta_0$. As $\lambda_1$ increases, the values of $\beta$ decrease, resulting in a more constrained model.

The Lasso Regression model is defined by the following optimization problem:
\begin{equation}
\hat{\beta}_{\text{LR}} = \arg\min_{\beta} \left\{ \sum_{i=1}^{n}(y_i - \sum_{j=1}^{p} x_{ij}\beta_j - \beta_0)^2 + \lambda_1 \sum_{j=1}^{p} |\beta_j| \right\}
\end{equation}

Here, $n$ is the number of observations, $p$ is the number of predictors, $y_i$ is the response at observation $i$, $x_{ij}$ is the $j$-th predictor at observation $i$, $\beta_j$ are the coefficients associated with each predictor, $\beta_0$ is the intercept, and $\lambda_1$ is the positive regularization parameter.

The term $\sum_{j=1}^{p} |\beta_j|$ in the objective function is the L1 norm of the coefficient vector, leading to a sparse solution by encouraging some coefficients to be exactly zero. As $\lambda_1$ increases, the penalty for non-zero coefficients strengthens, promoting sparsity in the model.


\subsection{Decision Tree}
Decision Tree (DT)\cite{navada2011overview} is a also widely used model in machine learning that operates by recursively partitioning the dataset into subsets based on the values of input features. The model is constructed in a hierarchical tree-like structure, where each internal node represents a decision based on a particular feature, and each leaf node corresponds to the predicted outcome.

Decision Trees are effective for both classification and regression tasks. The model is built by selecting the best feature at each node to split the data, optimizing for criteria such as Gini impurity for classification or mean squared error for regression.

The Decision Tree model can be expressed as follows:

\begin{equation}
\hat{y}_{\text{DT}} = \text{Tree}(X, \theta)
\end{equation}

Here, $\hat{y}_{\text{DT}}$ represents the predicted outcome, $X$ is the input feature vector, and $\theta$ represents the parameters learned during the training process. The tree structure is recursively defined through decisions at each node, leading to a final prediction at the leaf nodes.

Decision Trees are known for their interpretability and ability to capture complex relationships in the data. They are a fundamental building block in ensemble methods like Random Forests and Gradient Boosting.

Moreover, to further enhance the performance of the Random Forest model, certain parameters were set during the initialization of the RandomForestClassifier:

\begin{itemize}
\item \textbf{max\_depth = 5}: Restricts the maximum depth each tree in the forest can reach.
\item \textbf{min\_samples\_leaf = 7}: Specifies the minimum number of samples required in a leaf node.
\item \textbf{min\_samples\_split = 2}: Specifies that a node should have at least ten rows before it can be split.
\end{itemize}

\subsection{Random Forest}
Random Forest (RF)\cite{rigatti2017random} is a powerful ensemble learning method that combines the predictions of multiple decision trees. The ensemble nature of Random Forest makes it robust and effective for a variety of machine learning tasks. In Random Forest, each decision tree is constructed using a subset of the training data and a random subset of features.

The Random Forest model is formulated as follows:

\begin{equation}
\hat{Y}_{\text{RF}} = \frac{1}{B} \sum_{b=1}^{B} f_b(X)
\end{equation}

where $\hat{Y}_{\text{RF}}$ is the ensemble prediction, $B$ is the number of trees in the forest, and $f_b(X)$ is the prediction of the $b$-th decision tree.

The construction of each decision tree in the Random Forest involves:

1. Randomly selecting a subset of the training data with replacement (bootstrap samples).

2. Randomly selecting a subset of features at each split.

This process ensures that each decision tree in the Random Forest is built with a different subset of features, contributing to the overall diversity of the ensemble.

Moreover, to further enhance the performance of the Random Forest model, certain parameters were set during the initialization of the RandomForestClassifier:

\begin{itemize}
\item \textbf{max\_depth = 10}: Restricts the maximum depth each tree in the forest can reach.
\item \textbf{min\_samples\_leaf = 5}: Specifies the minimum number of samples required in a leaf node.
\item \textbf{min\_samples\_split = 10}: Specifies that a node should have at least ten rows before it can be split.
\item \textbf{n\_estimators = 100}: Specifies the number of trees in the Random Forest ensemble.
\end{itemize}

By iterating the model multiple times with these constraints and parameters, the Random Forest becomes a robust and high-performing ensemble classifier, capable of handling complex relationships in the data.

\section{LLM-Based House Price Prediction}

\subsection{Prompt-as-Prefix\cite{jin2023time} for House Price Prediction}

Prompting serves as a straightforward yet effective approach to task-specific activation of Large Language Models (LLM). However, directly translating house price data into natural language presents considerable challenges, hindering the creation of instruction-following datasets and the effective utilization of on-the-fly prompting without performance compromise. Inspired by recent advancements indicating that various data modalities can be seamlessly integrated as prefixes of prompts, we applied this concept to guide the transformation of house price data.

\subsubsection{Prompt-as-Prefix Construction}

We constructed our prompts by embedding key details about the house properties into the prompt's context. This includes data points such as location, type of house, area in square feet, and other relevant features. Our approach ensures that the LLM is provided with sufficient context to make accurate predictions.

\paragraph{Example Prompt:}
Below is an example of how the prompts were structured:

\begin{table}[h]
\caption{Example Prompt Structure}
\centering
\begin{tabular}{|l|p{0.35\textwidth}|}
\hline
\textbf{Field} & \textbf{Description} \\
\hline
\textbf{[Location]} & The house is located in [Location]. \\
\hline
\textbf{[Type and area]} & The property is a [house type], with an area of [area] square feet. \\
\hline
\textbf{[Features]} & It includes [number] bedrooms, [number] living rooms, [number] bathrooms, and features such as [air conditioner, heat, outdoorspace]. \\
\hline
\textbf{[Instruction]} & Predict the house price based on the above information. \\
\hline
\textbf{[Statistics]} & The training data includes houses with prices ranging from [min\_price] to [max\_price], with a median price of [median\_price]. The market trend is [upward/downward]. \\
\hline
\end{tabular}
\end{table}

\subsection{Forecasting Setups}

\subsubsection{Zero-shot Forecasting}
For the zero-shot scenario, the LLM was asked to predict house prices based solely on the provided prompts without any prior training data. This tests the model's inherent ability to understand and reason based on the contextual information provided in the prompt.

\subsubsection{Few-shot Forecasting}
\begin{itemize}
    \item \textbf{1-shot Forecasting:} Each prediction was based on one training data point, selected to be as similar as possible to the test data in terms of location, house type, and other features.
    \item \textbf{5-shot Forecasting:} Each prediction was based on five training data points, selected to be as similar as possible to the test data in terms of location, house type, and other features.
    \item \textbf{10-shot Forecasting:} Each prediction was based on ten training data points, again chosen to be closely related to the test data.
\end{itemize}

\paragraph{Selection Criteria:}
The training data for few-shot learning was selected with the goal of maintaining high relevance to the test data. Specifically, we ensured that the training samples were from the same geographical area and shared similar attributes, such as house type, area, and amenities.

\subsection{Robustness and Performance}

To evaluate the robustness of our approach, we examined whether variations in the prompt phrasing would significantly impact the performance. We experimented with different ways of presenting the house information, such as varying the order of details, using synonyms, and changing the format of the statistical information.

\subsubsection{Results}
Our findings indicated that the performance of the LLM was relatively stable across different phrasings of the prompt. Minor variations in wording did not significantly alter the predictions, demonstrating the robustness of the prompt-as-prefix approach in this context.

\paragraph{Performance Metrics:}
We measured the performance of our approach using Mean Squared Error (MSE), Mean Absolute Error (MAE), and R-Squared (R\textsuperscript{2}). The results are summarized in the table below:

\begin{table}[h]
\caption{Performance Metrics of LLM for House Price Prediction}
\centering
\begin{tabular}{lccc}
\hline
\textbf{Method} & \textbf{MSE} & \textbf{MAE} & \textbf{R-Squared} \\
\hline
ChatGPT (0-shot) & 9.47e+7 & 4.45e+3 & 0.46 \\
ChatGPT (1-shot) & 1.06e+8 & 4.67e+3 & 0.39 \\
ChatGPT (5-shot) & \textbf{6.09e+7} & \textbf{3.71e+3} & 0.65 \\
ChatGPT (10-shot) & 7.38e+7 & 3.85e+3 & \textbf{0.80} \\
\hline
\end{tabular}
\end{table}

\begin{itemize}
    \item \textbf{Zero-shot Forecasting:} Achieved satisfactory accuracy, leveraging the LLM's pre-existing knowledge and reasoning capabilities.
    \item \textbf{1-shot Forecasting:} Showed a slight decrease in performance compared to zero-shot, likely due to limited training data.
    \item \textbf{5-shot Forecasting:} Showed a notable improvement over zero-shot and 1-shot, with a significant reduction in Mean Squared Error (MSE) and Mean Absolute Error (MAE), and an increase in R-Squared (R\textsuperscript{2}).
    \item \textbf{10-shot Forecasting:} Further enhanced performance, achieving the highest R-Squared (R\textsuperscript{2}) value, indicating that additional relevant data points help in fine-tuning the model's predictions.
\end{itemize}

\subsection{Conclusion}

The prompt-as-prefix methodology effectively utilizes the LLM's capabilities to predict house prices, showing resilience to changes in prompt construction. This method offers a practical solution for integrating context-specific data into LLM prompts, facilitating accurate and reliable predictions in real-world scenarios.

\section{Results}

Numerous rounds of performance tuning were conducted to identify the optimal solutions for each model. This included both traditional machine learning models—such as Ridge Regression (RR), Lasso Regression (LR), Decision Tree (DT), and Random Forest Regression (RF)—and the Large Language Model (LLM) approach using ChatGPT. For the traditional models, extensive tuning was performed using the GridSearchCV function provided by scikit-learn. The LLM models were evaluated in 0-shot, 1-shot, 5-shot, and 10-shot scenarios to assess their performance. The outcomes achieved through this tuning and evaluation process are detailed in TABLE \uppercase\expandafter{\romannumeral4}. And Fig. 5 and Fig. 6 illustrate the performance of the different methods and the impact of varying the number of training shots on ChatGPT, respectively.

\begin{table}[h]
\centering
\caption{A comparison of different methods}
\begin{minipage}[t]{0.5\textwidth} 
\centering
\begin{tabular}{|l|ccc|}
\hline
 & \textbf{MSE} & \textbf{MAE} & \textbf{R Squared} \\ \hline
MLR  & 4.83e+7 & 3.42e+3 & \textbf{0.74} \\ 
RR  & 4.00e+7 & 3.40e+3 & 0.72\\ 
LR  & 3.98e+7 & 3.36e+3 & 0.72 \\ 
DT  & 3.88e+7 & 3.29e+3 & 0.73 \\ 
RF  & \textbf{3.71e+7} & \textbf{3.06e+3} & \textbf{0.74} \\ \hline
ChatGPT(0-shot) & 9.47e+7 & 4.45e+3 & 0.46 \\ 
ChatGPT(1-shot) & 1.06e+8 & 4.67e+3 & 0.39 \\ 
ChatGPT(5-shot) & \textbf{6.09e+7} & \textbf{3.71e+3} & 0.65 \\ 
ChatGPT(10-shot) & 7.38e+7 & 3.85e+3 & \textbf{0.80} \\ \hline
\end{tabular}
\end{minipage}
\begin{minipage}[t]{0.5\textwidth}
\centering
\footnotesize
\textbf{Notes:} 
\begin{itemize}
    \item MLR: Multiple Linear Regression
    \item RR: Ridge Regression
    \item LR: Lasso Regression
    \item DT: Decision Tree
    \item RF: Random Forest
    \item MSE: Mean Squared Error
    \item MAE: Mean Absolute Error
    \item R Squared: Coefficient of Determination
\end{itemize}
\end{minipage}
\end{table}

\begin{figure*}[htbp]
\centerline{\includegraphics[width=\linewidth]{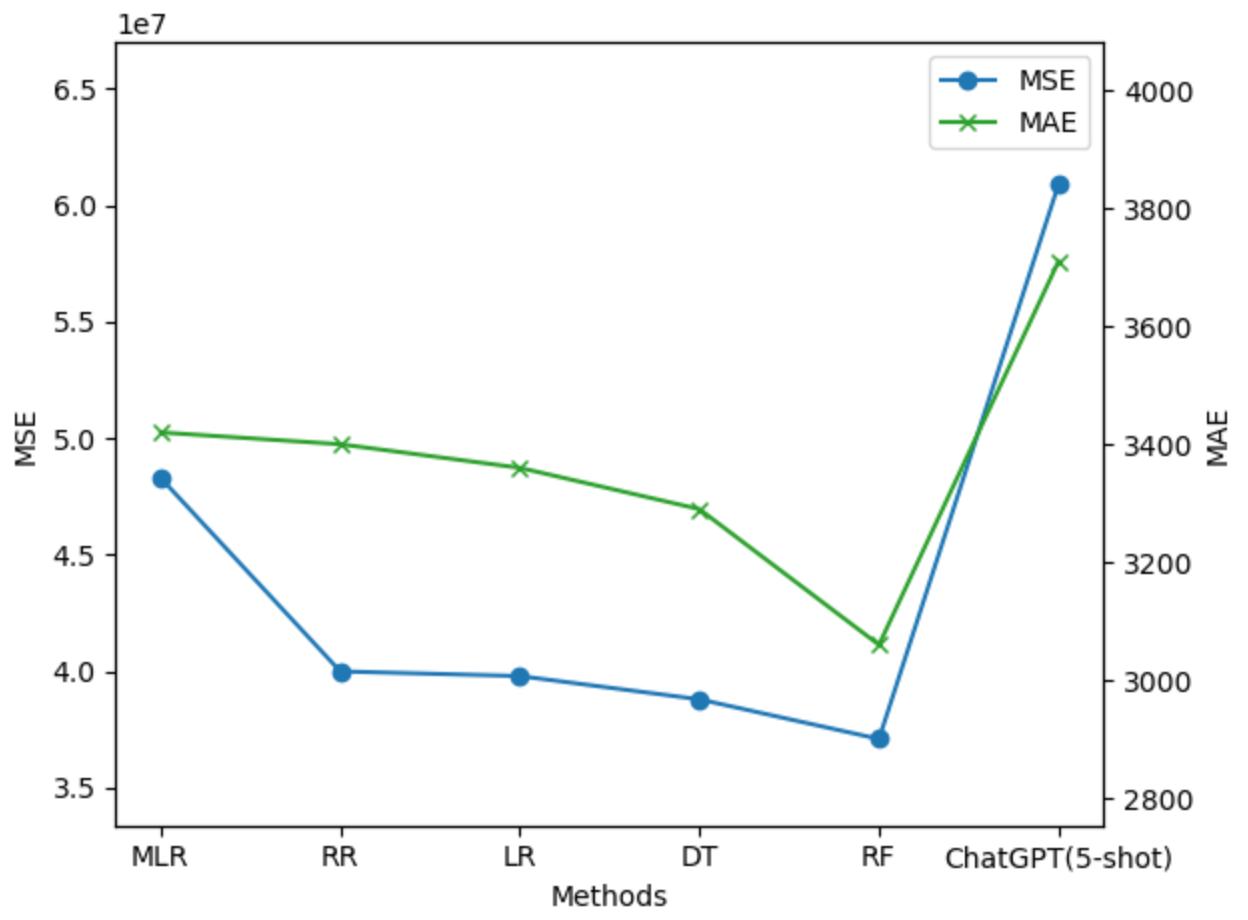}}
\caption{Comparison of different methods.}
\label{fig}
\end{figure*}

\begin{figure*}[htbp]
\centerline{\includegraphics[width=\linewidth]{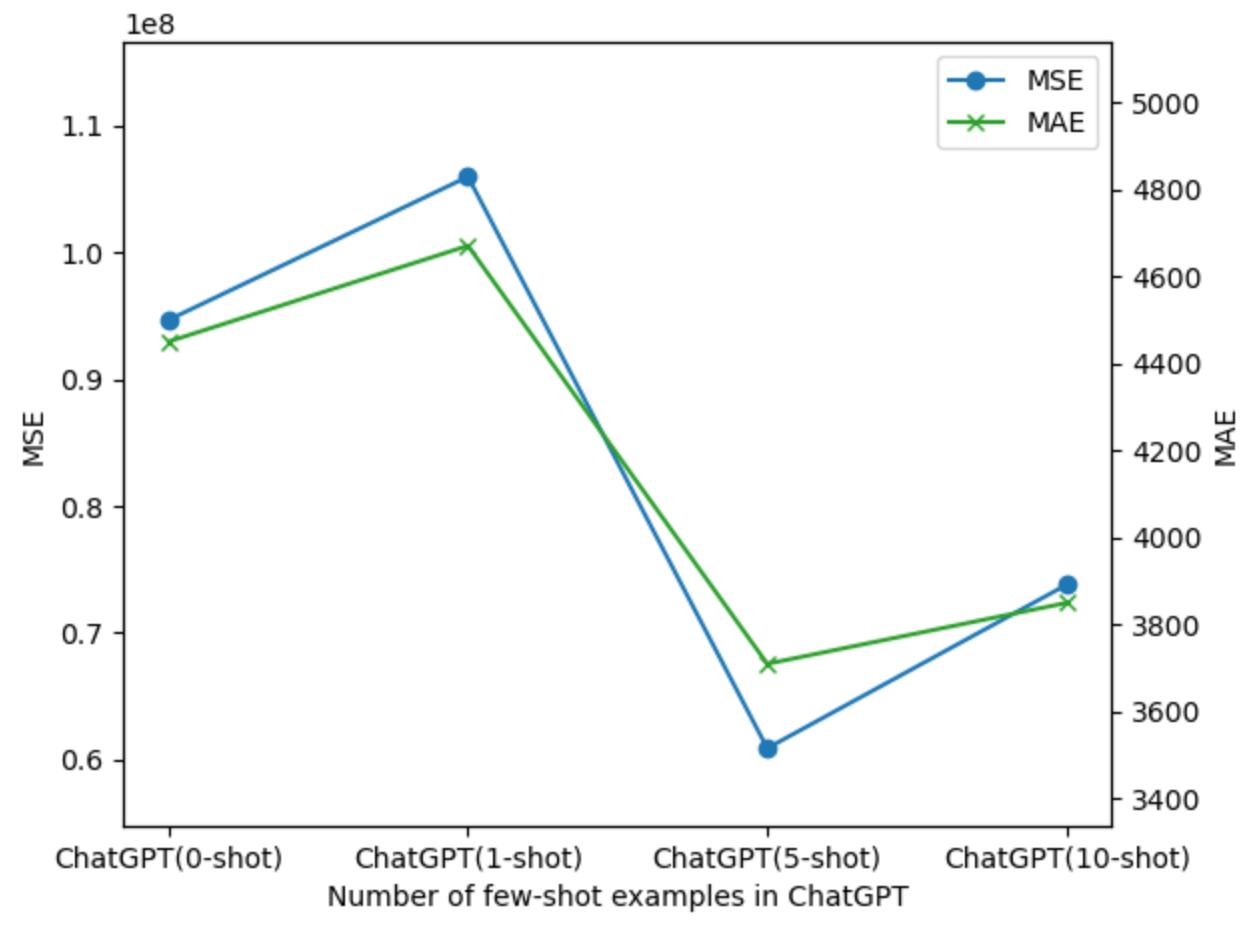}}
\caption{Comparison of different-shot examples in ChatGPT.}
\label{fig}
\end{figure*}

The analysis reveals that Random Forest (RF) stands out with the best performance, achieving the lowest Mean Squared Error (MSE) and Mean Absolute Error (MAE) among the considered regression models. This success can be attributed to the dataset's characteristics, with 49 out of 58 features being boolean values, making Random Forest well-suited for such conditions. Additionally, the R-squared values, a measure of the models' goodness of fit, indicate that Multiple Linear Regression (MLR) and Random Forest achieve the highest values at 0.74, indicating a strong correlation between predicted and actual values. However, it is acknowledged that Random Forest tends to be prone to overfitting, impacting its performance on unseen data. 

On the other hand, the Large Language Model (LLM) approach using ChatGPT demonstrates promising results, particularly with 5-shot and 10-shot forecasting. While the zero-shot and 1-shot performances lag behind traditional machine learning models, the 5-shot and 10-shot methods show significant improvements. The 10-shot method, in particular, achieves the highest R-Squared value of 0.80, surpassing all traditional machine learning models.

\section{Dissussion and Future Prospects}

Comparing the machine learning models with the LLM approach reveals some key insights:
\begin{itemize}
    \item Traditional machine learning models, especially Random Forest, perform well with structured data and exhibit strong predictive power.
    \item LLM, while initially less accurate in zero-shot and 1-shot scenarios, demonstrate substantial improvements with increased data shots. The 5-shot and 10-shot methods indicate that LLM can leverage additional context effectively to enhance their predictions.
    \item The highest R-Squared value achieved by ChatGPT (10-shot) suggests that with more refined and relevant training data, LLMs have the potential to surpass traditional models.
\end{itemize}

The results underscore the future potential of LLM in predictive modeling. With further advancements and fine-tuning, LLM could become formidable tools in various domains, offering flexibility and adaptability that traditional models may lack. The ability of LLM to handle unstructured data and generate insights from minimal input makes them highly promising for future applications.

In conclusion, while traditional machine learning models currently hold strong performance in structured data scenarios, the emerging capabilities of LLM like ChatGPT highlight a significant shift towards more versatile and adaptive predictive modeling techniques.

\bibliographystyle{unsrt}
\bibliography{reference}

\end{document}